\documentclass[runningheads,a4paper]{llncs}

\usepackage{amssymb}
\usepackage{graphicx}
\usepackage{siunitx}
\usepackage{mathtools}
\usepackage{threeparttable}
\usepackage{booktabs}
\usepackage{tabularx}
\newcolumntype{C}[1]{>{\centering\arraybackslash}p{#1}}
\newcolumntype{L}[1]{>{\raggedright\arraybackslash}p{#1}}
\usepackage{multicol}

\usepackage{url}
\newcommand{\keywords}[1]{\par\addvspace\baselineskip
\noindent\keywordname\enspace\ignorespaces#1}
\usepackage{subcaption}
\begin{document}

\mainmatter

\title{Fast Vessel Segmentation and Tracking in Ultra High-Frequency Ultrasound Images}

\titlerunning{Fast Vessel Segmentation and Tracking in UHFUS}


\author{Tejas Sudharshan Mathai\inst{1}, Lingbo Jin\inst{2}, Vijay Gorantla\inst{3}, John Galeotti\inst{1} } %

\authorrunning{T.S. Mathai et al.}

\institute{The Robotics Institute, Carnegie Mellon University, Pittsburgh PA 15213, USA \and
Department of ECE, Carnegie Mellon University, Pittsburgh PA 15213, USA \and 
Department of Surgery, Wake Forest Institute for Regenerative Medicine, Winston-Salem NC 27101, USA}

\maketitle

\begin{abstract}
Ultra High Frequency Ultrasound (UHFUS) enables the visualization of highly deformable small and medium vessels in the hand. Intricate vessel-based measurements, such as intimal wall thickness and vessel wall compliance, require sub-millimeter vessel tracking between B-scans. Our fast GPU-based approach combines the advantages of local phase analysis, a distance-regularized level set, and an Extended Kalman Filter (EKF), to rapidly segment and track the deforming vessel contour. We validated on 35 UHFUS sequences of vessels in the hand, and we show the transferability of the approach to 5 more diverse datasets acquired by a traditional High Frequency Ultrasound (HFUS) machine. To the best of our knowledge, this is the first algorithm capable of rapidly segmenting and tracking deformable vessel contours in 2D UHFUS images. It is also the fastest and most accurate system for 2D HFUS images.  
\keywords{Ultrasound, Vasculature, Segmentation, Tracking}
\end{abstract}

\section{Introduction}
\label{intro}
Ultra High Frequency Ultrasound (UHFUS) is a new advancement in non-invasive imaging, capable of operating above 50 MHz and resolving structures less than $0.03mm$. Potential clinical applications include vascular measurements for surgical procedures and disease diagnosis, with such measures including intimal wall thickness and variations in atherosclerotic plaque buildup \cite{Gorantla11}. It can be used to monitor hand transplant recipients \cite{Gorantla11}, for whom the gold standard diagnosis using invasive histopathology is not practical due to suppressed immune systems \cite{Gorantla11}. However, UHFUS can only image through $\sim$1cm of tissue. Vessels can be visualized at this depth (see Fig. \ref{fig:fig1}(a)), in contrast to skeletal structures, which are too deep to be imaged. When compared against traditional high frequency ultrasound (HFUS) (see Fig. \ref{fig:fig1}(b)), substantially increased speckle noise is encountered with UHFUS at such shallow depths. The vessel measurements have naturally occurring sub-millimeter(mm) variations along their length, and sub-mm displacements of the probe confound comparisons across time. Our motivation is that vessel tracking across B-scans with sub-mm precision should enable consistent comparisons. \textit{In this work, the primary medical image computing (MIC) goal is the fast sub-mm 2D vessel contour localization.}

Traditional ultrasound based real-time vessel tracking has been researched before \cite{Abolmaesumi00,Guerrero07,Wang09,Smistad16}. However, when tested on UHFUS images, these gradient-based edge detection approaches failed to detect and track the vessel boundaries in the presence of higher speckle noise. Furthermore, precise delineation of the deforming vessel is required for vessel-based measurements, whereas prior approaches \cite{Abolmaesumi00,Guerrero07,Wang09,Smistad16} modeled the vessel as an ellipse without accounting for the deforming vessel contour. A recent approach in \cite{Smistad16} was designed for a specific imaging setting of 55$\%$ maximum gain, but when applied to UHFUS sequences, it completely failed to track vessels regardless of gain settings (see Fig. \ref{fig:fig1}(c)). A recent level-set based approach \cite{Chaniot16} designed for HFUS images ran slowly at 0.5 seconds per image. 

In this paper, a fast GPU-based approach is presented to segment and track the deforming vessel contour in UHFUS images. It combines the robust edge detection capability of local phase analysis, with a distance regularized level set to accurately capture the vessel contour, and an efficient Extended Kalman Filter (EKF) to track the vessel. Validation on 35 UHFUS sequences showed that it successfully segmented and tracked vessels undergoing dynamic compression. Our algorithm achieved a maximum Hausdorff distance error of 0.135mm, which was 6$\times$ smaller than the smallest vessel diameter of 0.81mm. It also generalized to datasets acquired with different imaging settings and from a HFUS imaging system, with errors $\sim$2$\times$ smaller than the state-of-the-art for HFUS \cite{Chaniot16}. 

\noindent
\textbf{Contribution.} 1) We present the first system capable of rapidly segmenting and tracking a vessel contour in UHFUS images, and we demonstrate its high speed performance ($\geq$52 FPS). 2) We demonstrate the generality of our approach by applying it to datasets acquired from a traditional HFUS machine, and show that it is faster than the state-of-the-art approach for HFUS.


\section{Methods}
\label{methods}

\subsection{Data Acquisition}
\label{dataAcq}
The Visualsonics Vevo 2100 UHFUS machine (Fujifilm, Canada) and a 50 MHz transducer (bandwidth extendable to 70 MHz) was used to acquire freehand ultrasound volumes. This UHFUS system has a physical resolution of 30\SI{}{\micro\meter}, and the pixel pitch is 11.6\SI{}{\micro\meter} between pixel centers. 35 deidentified UHFUS sequences were acquired over a wide range of gain values (40-70 dB), with the maximum gain value setting being 70 dB. The sequences contained a wide range of motions with the probe, such as longitudinal scanning, out-of-plane tissue deformation, beating vessel visualization, etc. Fig. \ref{fig:fig1}(a) shows an example ultrasound image of the proper palmar digital artery acquired with the UHFUS system. Each sequence consisted of 100 2D B-scans with dimensions of 832$\times$512 pixels. To show the generality of our approach, 5 additional sequences were acquired from a traditional HFUS machine (Diasus, Dynamic Imaging, UK) using a 10-22 MHz transducer. The pixel resolution for the HFUS machine was 92.5\SI{}{\micro\meter}, and each sequence consisted of 250 2D B-scans of dimensions 280$\times$534 pixels. 

\subsection{Noise Reduction and Clustering}
\label{nrc}
\begin{figure}[!h]
\centering
\begin{subfigure}[b]{0.19\columnwidth}
\vspace*{\fill}
  \centering
  \includegraphics[width=\columnwidth,height=1.8cm]{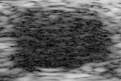}
  \centerline{(a)} 
  \label{fig:1a}
  \includegraphics[width=\columnwidth,height=1.8cm]{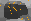}
  \centerline{(f)}
  \label{fig:1b}
\end{subfigure} 
\begin{subfigure}[b]{0.19\columnwidth}
\vspace*{\fill}
  \centering
  \includegraphics[width=\columnwidth,height=1.8cm]{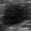}
  \centerline{(b)} 
  \label{fig:1c}
  \includegraphics[width=\columnwidth,height=1.8cm]{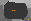}
  \centerline{(g)} 
  \label{fig:1d}
\end{subfigure} 
\begin{subfigure}[b]{0.19\columnwidth}
\vspace*{\fill}
  \centering
  \includegraphics[width=\columnwidth,height=1.8cm]{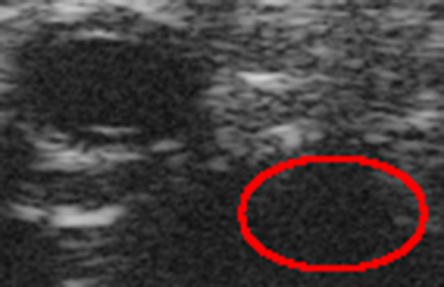}
  \centerline{(c)} 
  \label{fig:1e}
  \includegraphics[width=\columnwidth,height=1.8cm]{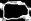}
  \centerline{(h)} 
  \label{fig:1f}
\end{subfigure} 
\begin{subfigure}[b]{0.19\columnwidth}
\vspace*{\fill}
  \centering
  \includegraphics[width=\columnwidth,height=1.8cm]{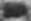}
  \centerline{(d)} 
  \label{fig:1g}
  \includegraphics[width=\columnwidth,height=1.8cm]{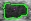}
  \centerline{(i)} 
  \label{fig:1h}
\end{subfigure} 
\begin{subfigure}[b]{0.19\columnwidth}
\vspace*{\fill}
  \centering
  \includegraphics[width=\columnwidth,height=1.8cm]{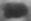}
  \centerline{(e)} 
  \label{fig:1i}
  \includegraphics[width=\columnwidth,height=1.8cm]{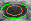}
  \centerline{(j)} 
  \label{fig:1j}
\end{subfigure}
\caption{Vessel imaged using (a) UHFUS, (b) HFUS; (c) Failed vessel detection result (red ellipse) of algorithm in \cite{Smistad16} on an UHFUS image; (d) Downsampled image; (e) Bilateral filtered image (${I}_{\textnormal{B}}$); (f) With a kernel size $3 \times 3$, pixels in ${I}_{\textnormal{B}}$ are clustered into homogeneous patches in ${I}_{\textnormal{C}}$, each with its own root (orange points); (g) ${I}_{\textnormal{C}}$ generated  with $7 \times 7$ kernel; (h) Feature Asymmetry map (${I}_{\textnormal{FA}}$); (i) Initial boundary locations (green points) estimated from ${I}_{\textnormal{FA}}$ using the tracked point $\mathbf{s}^{t}$ (magenta); (j) Ellipse (green) fitted to green points in (i), and then shrunk (brown ellipse) to initialize the level set evolution.}
\label{fig:fig1}
\end{figure}

\textbf{Noise Reduction.} In contrast to traditional HFUS, speckle noise is greater in UHFUS as seen in Figs. \ref{fig:fig1}(a) and \ref{fig:fig1}(b). To mitigate the effects of speckle during segmentation and speed up computation, the UHFUS B-scans were first downsampled by a factor of 4 in each dimension (see Fig. \ref{fig:fig1}(d)). Next, a bilateral filter \cite{Tomasi1998} of size 5$\times$5 pixels was applied to the downsampled image to smooth the small amplitude noise (see Fig. \ref{fig:fig1}(e)), while preserving vessel boundaries that are crucial to our segmentation. The bilateral filtered image is represented by ${I}_{\textnormal{B}}$.

\noindent
\textbf{Clustering.} The approach published in \cite{Stetten13}, which has also shown applicability to MRI images, was used to produce an image ${I}_{\textnormal{C}}$, where the pixels in ${I}_{\textnormal{B}}$ were clustered into homogeneous patches (see Figs. \ref{fig:fig1}(f) and \ref{fig:fig1}(g)). Each pixel in ${I}_{\textnormal{C}}$ can be represented by two elements: the mean intensity of the patch that it belongs to, and a cluster/patch center (root). For each pixel in ${I}_{\textnormal{B}}$, the mean intensity and variance is found in a circular neighborhood, whose size varies depending on the size of the vessel. For small vessels in UHFUS images ($\leq$70 pixel diameter or 0.81mm), the neighborhood size was 3$\times$3 pixels, while it was 7$\times$7 pixels for larger vessels ($>$70 pixels). Each patch root in ${I}_{\textnormal{C}}$ has the lowest local variance amongst all the members of the same patch \cite{Stetten13}. Roots in ${I}_{\textnormal{C}}$ were used solely as seeds to track vessels over sequential B-scans. As seen in Figs. \ref{fig:fig1}(f) and \ref{fig:fig1}(g), increasing the neighborhood size reduces the number of roots that can be tracked, which can cause tracking failure when large motion occurs. 



\subsection{Local Phase Analysis}
\label{phase}
Vessel boundaries in ${I}_{\textnormal{B}}$ were highlighted using a Cauchy filter, which has been shown to be better than a Log-Gabor filter at detecting edges in ultrasound \cite{Belaid11}. We denote the spatial intensity value at a location $\mathbf{x}$=$[{x}\enskip{y}]^{T}$ in the image ${I}_{\textnormal{B}}$ by ${I}_{\textnormal{B}}(\mathbf{x})$. After applying a 2D Fourier transform, the corresponding 2D frequency domain value is ${F(\mathbf{w})}$, where $\mathbf{w} = [{w}_{1}\enskip{w}_{2}]^{T}$. The Cauchy filter ${C}(\mathbf{w})$ applied to ${F(\mathbf{w})}$ is represented as: 
\begin{equation}          
			\label{eq1}
           \begin{aligned}  
           C(\mathbf{w}) = {\Vert \mathbf{w} \Vert_2^{u}} \exp \left( -{w}_{o} \Vert \mathbf{w} \Vert_2 \right),\qquad  {u} \geq {1}                 
          \end{aligned}
\end{equation}
\noindent
where ${u}$ is a scaling parameter, and ${w}_{o}$ is the center frequency. We chose the same optimal parameter values suggested in \cite{Belaid11}: ${w}_{o}$=10, and $u$=1. Filtering ${F(\mathbf{w})}$ with ${C}(\mathbf{w})$ yielded the monogenic signal, from which the feature asymmetry map (${I}_{\textnormal{FA}}$) \cite{Belaid11} was obtained (see Fig. \ref{fig:fig1}(h)). Pixel values in ${I}_{\textnormal{FA}}$ range between [0, 1]. 

\subsection{Vessel Segmentation and Tracking}
\label{vsc}

\textbf{Initialization.} As in \cite{Guerrero07,Wang09}, we manually initialize our system by clicking a point inside the vessel lumen in the first B-scan of a sequence. This pixel location is stored as a seed, denoted by ${\mathbf{s}}^{0}$ at time $t$=0, to segment the vessel boundary in the first B-scan, and initialize the vessel lumen tracking in subsequent B-scans. 

\noindent
\textbf{Initial Boundary Segmentation.} $N$ = 360 radial lines of maximum search length $M$ = 100, which corresponds to the largest observed vessel diameter, stem out from ${\mathbf{s}}^{0}$ to find the vessel boundaries in ${I}_{\textnormal{FA}}$. The first local maximum on each radial line is included in a set $\mathcal{I}$ as an initial boundary point (see Fig. \ref{fig:fig1}(i)). 
 
\begin{figure}[!h]
\centering
\begin{subfigure}[b]{0.2\columnwidth}
\vspace*{\fill}
  \centering
  \includegraphics[width=\columnwidth,height=1.8cm]{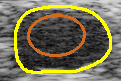}
  \centerline{(a)}
  \label{fig:2a}
\end{subfigure}\hfill
\begin{subfigure}[b]{0.19\columnwidth}
\vspace*{\fill}
  \centering
  \includegraphics[width=\columnwidth,height=1.8cm]{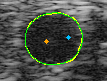}
  \centerline{(b)}
  \label{fig:2b}
\end{subfigure}\hfill
\begin{subfigure}[b]{0.19\columnwidth}
\vspace*{\fill}
  \centering
  \includegraphics[width=\columnwidth,height=1.8cm]{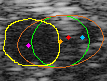}
  \centerline{(c)}
  \label{fig:2c}
\end{subfigure}\hfill
\begin{subfigure}[b]{0.19\columnwidth}
\vspace*{\fill}
  \centering
  \includegraphics[width=\columnwidth,height=1.8cm]{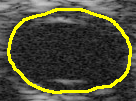}
  \centerline{(d)}
  \label{fig:2d}
\end{subfigure}\hfill
\begin{subfigure}[b]{0.19\columnwidth}
\vspace*{\fill}
  \centering
  \includegraphics[width=\columnwidth,height=1.8cm]{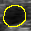}
  \centerline{(e)}
  \label{fig:2e}
\end{subfigure}
\caption{(a) Refined segmentation (yellow contour) evolved from initial LSF (brown ellipse); Tracking under large motion - (b) In frame 87, ${\mathbf{s}}_{\textnormal{ekf}}^{87}$ (blue) chosen over ${\mathbf{s}}_{\textnormal{c}}^{87}$ (orange) to segment vessel (yellow contour), which is then fitted with an ellipse (green); (c) In frame 88, the EKF prediction ${\mathbf{s}}_{\textnormal{ekf}}^{88}$ (red) is ignored as Eq. $\eqref{eq8}$ is not satisfied. Instead, ${\mathbf{s}}_{\textnormal{c}}^{88}$ (magenta) is chosen as it falls under the elliptical neighborhood (brown) of ${\mathbf{s}}_{\textnormal{c}}^{87}$ (orange); (d) Successful contour segmentation (Adventitia) of UHFUS image in Fig \ref{fig:fig1}(c); (e) Successful segmentation of vessel in HFUS image shown in Fig \ref{fig:fig1}(b).}
\label{fig:fig2}
\end{figure}

\noindent
\textbf{Segmentation Refinement.} A rough estimate of the semi-major and semi-minor vessel axes was determined by fitting an ellipse \cite{Fitzgibbon95} to the initial boundary locations in $\mathcal{I}$. Next, the estimated values were shrunk by $75\%$, and used to initialize an elliptical binary level set function (LSF) ${\phi}_{o}$ (see Fig. \ref{fig:fig1}(j)) in a narrowband distance regularized level set evolution (DRLSE) \cite{Li10} framework. As the LSF initialization is close to the true boundaries, the DRLSE formulation allows quick propagation of LSF to the desired vessel locations $\mathcal{D}$ (see Fig. \ref{fig:fig2}(a)) with a large timestep $\Delta{\tau}$ \cite{Li10}. The DRLSE framework minimizes an energy functional $\mathcal{E}(\phi)$ \cite{Li10} using the gradient defined in Eq. $\eqref{eq3}$. $\mu$, $\lambda$, $\epsilon$, and $\alpha$ are constants, $g$ is the same edge indicator function used in \cite{Li10}, and ${\delta}_{\epsilon}$ and ${d}_{p}$ are first order derivatives of the Heaviside function and the double-well potential respectively. The parameters used in all datasets were: $\Delta{\tau}=10, \mu=0.2, \lambda=1, \alpha=-1, \epsilon=1$ for a total of $15$ iterations. 
\begin{equation}          
			\label{eq3}
           \begin{aligned}  
           \frac{\partial\phi}{\partial\tau} = 
           \mu\text{div}({d}_{p} (|\nabla\phi|) \nabla\phi ) \, + \, 
           \lambda {\delta}_{\epsilon}(\phi) \text{div}\bigg(g \frac{\nabla\phi}{|\nabla\phi|}\bigg) \, + \, 
           \alpha {g} {\delta}_{\epsilon}(\phi)
          \end{aligned}
\end{equation}

\noindent
\textbf{Vessel Tracking.} To update the vessel lumen position ${\mathbf{s}}^{t}$ at time $t$ to ${\mathbf{s}}^{t+1}$ at time $t+1$, two new potential seeds are found, from which one is chosen. The first seed is found using an EKF \cite{Smistad16,Kalman60}. The second seed is found using ${I}_{\textnormal{C}}$, and it is needed in case the EKF fails to track the vessel lumen due to abrupt motion. The EKF tracks a state vector defined by: $\mathbf{x}^{t} = [c_x^{t}, c_y^{t}, a^{t}, b^{t}]$, where ${\mathbf{s}}_{\textnormal{ekf}}^{t}$=$[c_x^{t}, c_y^{t}]$ is the EKF-tracked vessel lumen location and $[a^{t}, b^{t}]$ are the tracked semi-major and semi-minor vessel axes respectively. Instead of tracking all locations in $\mathcal{D}$, it is computationally efficient to track $\mathbf{x}^{t}$, whose elements are estimated by fitting an ellipse once again to the locations in $\mathcal{D}$ (see Fig. \ref{fig:fig2}(b)). The EKF projects the current state $\mathbf{x}^{t}$ at time $t$ to the next state $\mathbf{x}^{t+1}$ at time $t$+1 using the motion model in \cite{Smistad16}, which uses two state transition matrices $A_1, A_2$, the covariance error matrix $P$, and the process-noise covariance matrix $Q$. These matrices are initialized in Eqs. $(\ref{eq4})$-$(\ref{eq7})$. 
\begin{align}
	A_1 \quad =& \quad diag([1.5, 1.5, 1.5, 1.5]) \label{eq4}\\
	A_2 \quad =& \quad diag([-0.5, -0.5, -0.5, -0.5]) \label{eq5}\\
	P \quad =& \quad diag([1000, 1000, 1000, 1000]) \label{eq6}\\ 
  	Q \quad =& \quad diag([0.001, 0.001, 0.001, 0.001]) \label{eq7}           
\end{align}


The second seed was found using the clustering result. At ${\mathbf{s}}_{\textnormal{c}}^{t}$ in the clustered image ${I}_{\textnormal{C}}^{t+1}$ at time $t$+1, the EKF tracked axes $[a^{t+1}, b^{t+1}]$ were used to find the neighboring roots of ${\mathbf{s}_{\textnormal{c}}^{t}}$ in an elliptical region of size $[{1.5{a}^{t+1}}, {b}^{t+1}]$ pixels. Amongst these roots, the root ${\mathbf{s}}_{\textnormal{c}}^{t+1}$, which has the lowest mean pixel intensity representing a patch in the vessel lumen, is chosen. By using the elliptical neighborhood derived from the EKF state, ${\mathbf{s}}_{\textnormal{c}}^{t}$ is tracked in subsequent frames (see Fig. \ref{fig:fig2}(c)). The elliptical region is robust to vessel compression, which enlarges the vessel horizontally.


The EKF prediction is sufficient for tracking during slow longitudinal scanning or still imaging as ${\mathbf{s}}_{\textnormal{ekf}}^{t+1}$ and ${\mathbf{s}}_{\textnormal{c}}^{t+1}$ lie close to each other. However, when large motion was encountered, the EKF incorrectly predicted the vessel location (see Fig \ref{fig:fig2}(c)) as it corrected motion, thereby leading to tracking failure. To mitigate tracking failure during large vessel motion, ${\mathbf{s}}_{\textnormal{ekf}}^{t+1}$ was ignored, and ${\mathbf{s}}_{\textnormal{c}}^{t+1}$ was updated as the new tracking seed according to the rule in Eq. $\eqref{eq8}$: 

\begin{align}
    {\mathbf{s}}^{t+1} \quad = & \quad 
    \begin{cases}
    {\mathbf{s}}_{\textnormal{c}}^{t+1} & if \quad \Vert {\mathbf{s}}_{\textnormal{ekf}}^{t+1} - {\mathbf{s}}_{\textnormal{c}}^{t+1} \Vert_2 > a^{t+1}\\
    {\mathbf{s}}_{\textnormal{ekf}}^{t+1} & \quad otherwise \label{eq8}
    \end{cases}     
\end{align}
\section{Results and Discussion}
\textbf{Metrics.} Segmentation accuracy of the proposed approach was evaluated by comparing the contour segmentations against the annotations of two graders. All images in all datasets were annotated by two graders. Tracking was deemed successful if the vessel was segmented in all B-scans of a sequence. Considering the set of ground truth contour points as $G$ and the segmented contour points as $S$, the following metrics were calculated as defined in Eqs. \eqref{eq9}-\eqref{eq12}: 1) \textit{Dice Similarity Coefficient} (DSC), 2) \textit{Hausdorff Distance} (H) in millimeters, 3) \textit{Definite False Positive and Negative Distances} (DFPD, DFND). The latter represent weighted distances of false positives and negatives to the true annotation. Let ${I}_{\textnormal{G}}$ and ${I}_{\textnormal{S}}$ be binary images containing 1 on and inside the area covered by $G$ and $S$ respectively, and 0 elsewhere. The Euclidean Distance Transform (EDT) is computed for ${I}_{\textnormal{G}}$ and its inverse ${I}_{\textnormal{G}}^{\textnormal{Inv}}$ \cite{Maurer03}. DFPD and DFND are estimated from the element-wise product of ${I}_{\textnormal{S}}$ with EDT(${I}_{\textnormal{G}}$) and EDT(${I}_{\textnormal{G}}^{\textnormal{Inv}}$) respectively $\eqref{eq11}$-$\eqref{eq12}$. $d(i,G,S)$ is the distance from contour point $i$ in $G$ to the closest point in $S$. Inter-grader annotation variability was also measured. 
\begin{align}
  	\textnormal{DSC} \quad =& \quad \frac{2\lvert G \cap S \rvert}{\lvert G \rvert + \lvert S \rvert} \label{eq9}\\
	\textnormal{H} \quad =& \quad \max\Big(\underset{i \in [1,|G|]}{\max} d(i,G,S), \underset{j \in [1,|S|]}{\max} d(j,S,G) \Big) \label{eq10} \\
	\textnormal{DFPD} \quad =& \quad \log\Big(\Vert \textnormal{EDT}({I}_{\textnormal{G}}) \circ {I}_{\textnormal{S}} \Vert_1\Big) \label{eq11}\\
	\text{DFND} \quad =& \quad \log\Big(\Vert \textnormal{EDT}({I}_{\textnormal{G}}^{\textnormal{Inv}}) \circ {I}_{\textnormal{S}} \Vert_1\Big) \label{eq12}
\end{align}
\begin{figure}[!h]
\centering
\begin{subfigure}[b]{0.25\columnwidth}
\vspace*{\fill}
  \centering
  \includegraphics[width=\columnwidth,height=2.1cm]{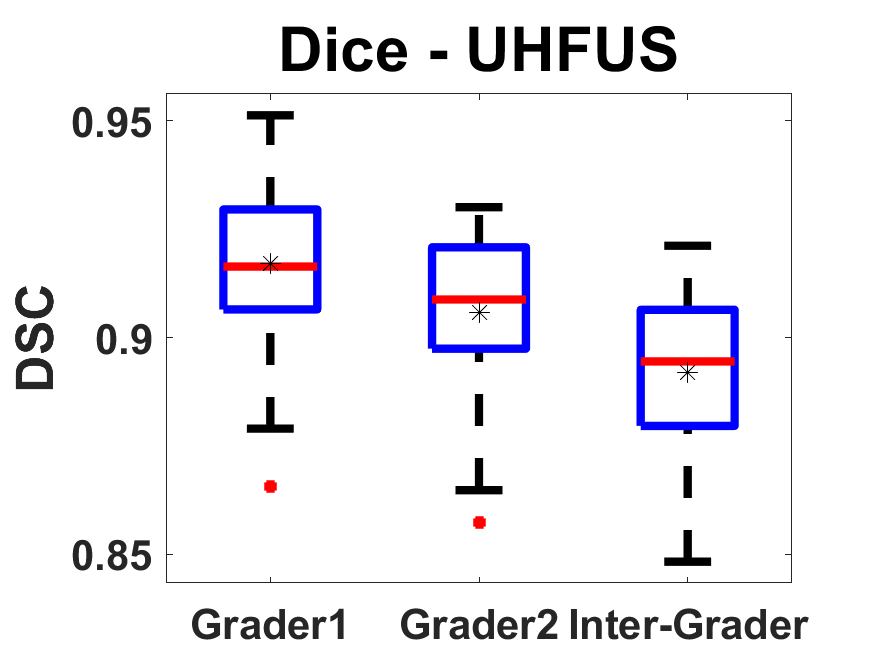}
  \centerline{(a)}  
  \includegraphics[width=\columnwidth,height=2.1cm]{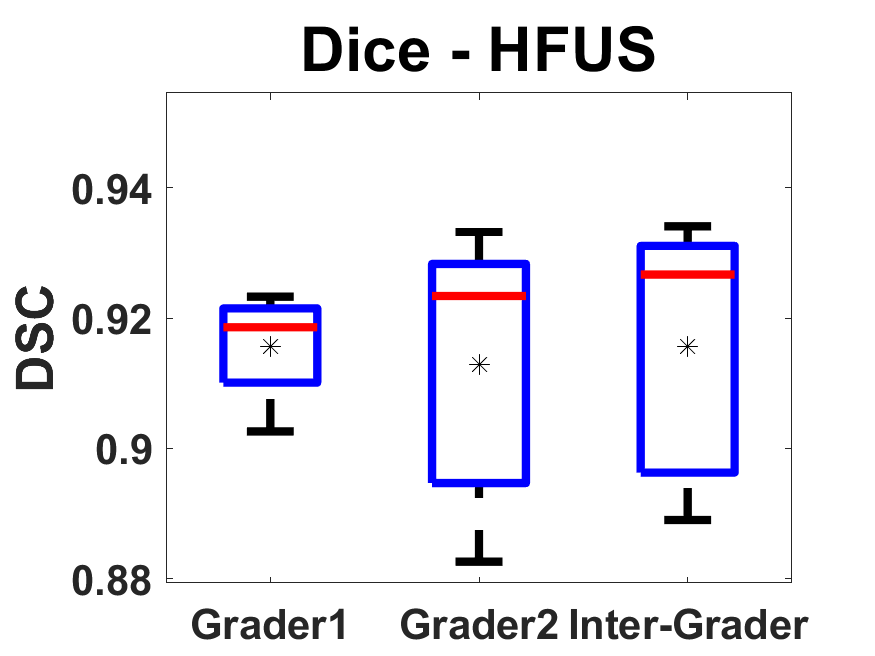}
  \centerline{(e)}
\end{subfigure}\hfill
\begin{subfigure}[b]{0.25\columnwidth}
\vspace*{\fill}
  \centering
  \includegraphics[width=\columnwidth,height=2.1cm]{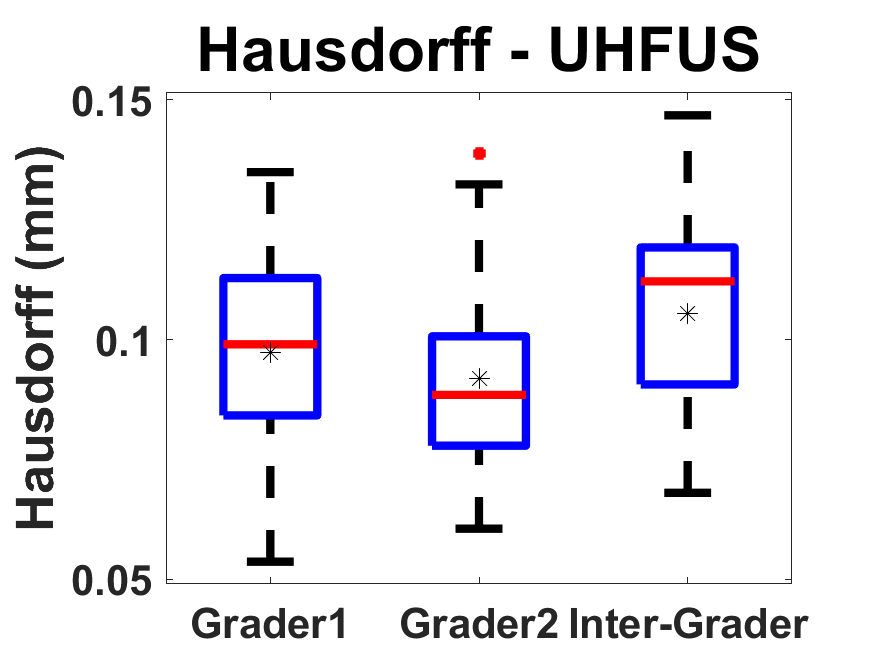}
  \centerline{(b)}
  \includegraphics[width=\columnwidth,height=2.1cm]{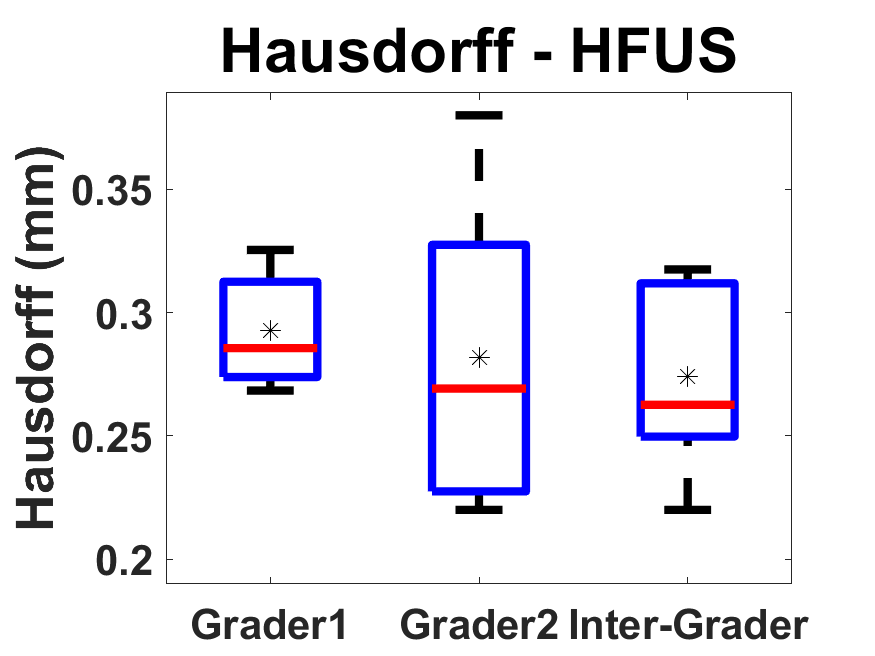}
  \centerline{(f)}
\end{subfigure}\hfill
\begin{subfigure}[b]{0.25\columnwidth}
\vspace*{\fill}
  \centering
  \includegraphics[width=\columnwidth,height=2.1cm]{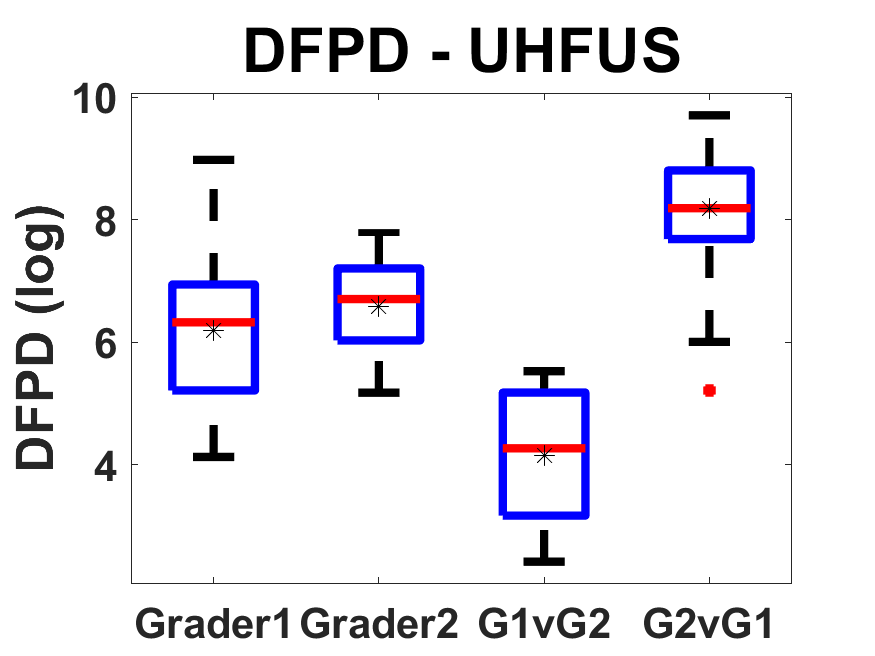}
  \centerline{(c)}
  \includegraphics[width=\columnwidth,height=2.1cm]{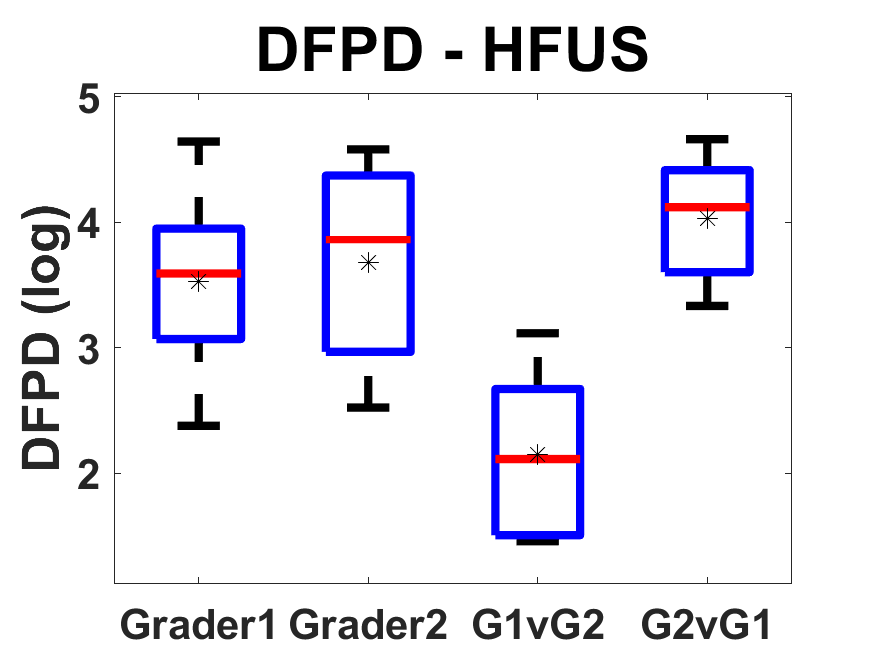}
  \centerline{(g)}
\end{subfigure}\hfill
\begin{subfigure}[b]{0.25\columnwidth}
\vspace*{\fill}
  \centering
  \includegraphics[width=\columnwidth,height=2.1cm]{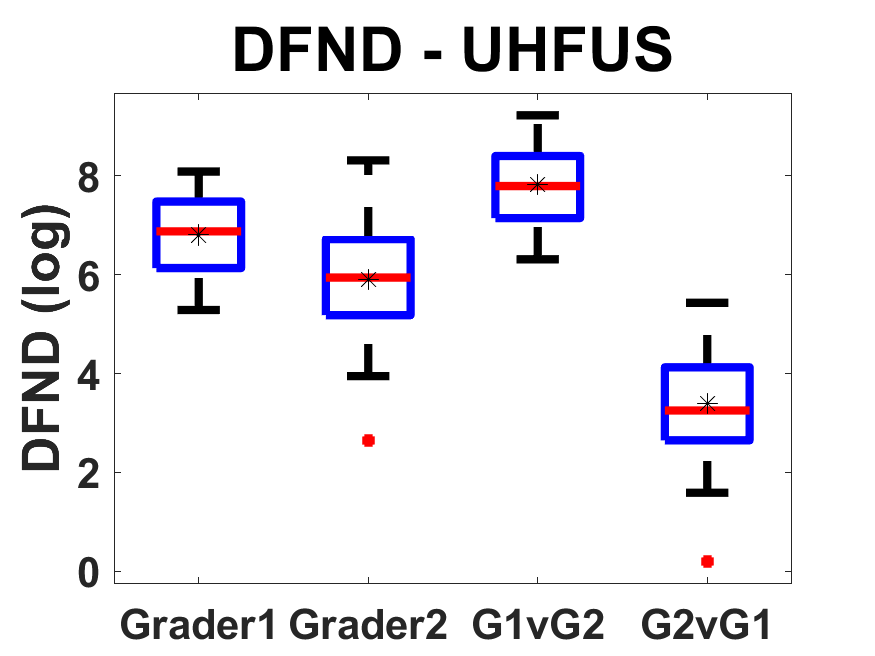}
  \centerline{(d)}
  \includegraphics[width=\columnwidth,height=2.1cm]{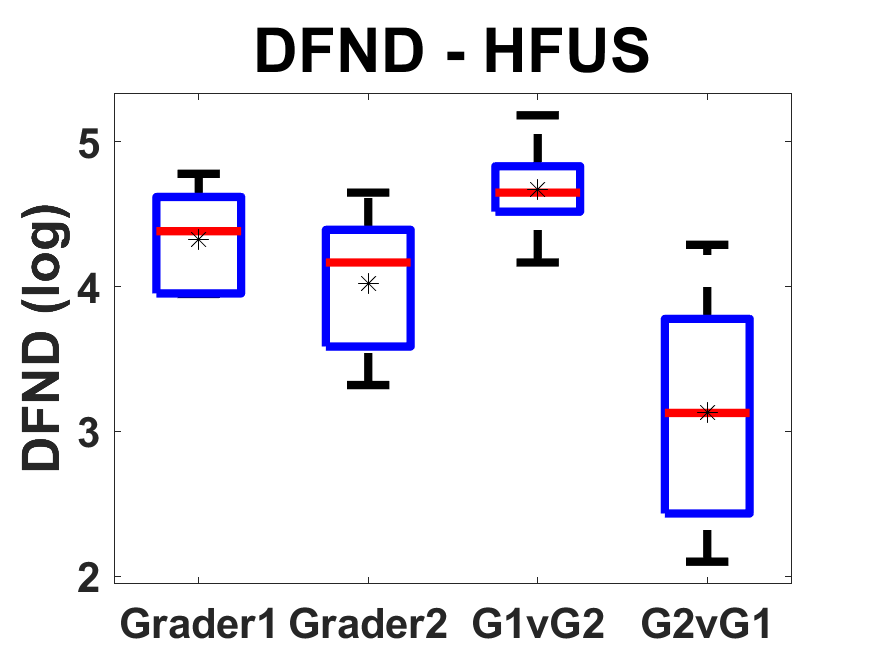}
  \centerline{(h)}
\end{subfigure}\hfill
\caption{Quantitative segmentation and tracking accuracy metrics for 35 UHFUS (top row) and 5 HFUS (bottom row) sequences respectively. The black * in each box plot represents the mean value of the metric. The terms 'G1vG2' and 'G2vG1' in Fig. \ref{fig:fig3} represent the inter-grader annotation variability when grader 2 annotation was considered the ground truth, and vice versa.}
\label{fig:fig3}
\end{figure}

\noindent
\textbf{UHFUS Results.} We ran our algorithm on 35 UHFUS sequences (100 images each), and the corresponding results are shown in Figs. \ref{fig:fig3}(a)-\ref{fig:fig3}(d). The two graders varied in their estimation of the vessel boundary locations in UHFUS images due to the speckle noise obscuring the precise location of the vessel edges, as shown in the inter-grader Dice score in Fig. \ref{fig:fig3}(a), inter-grader Hausdorff distance in Fig. \ref{fig:fig3}(b), and inter-grader variation between Figs. \ref{fig:fig3}(c) and \ref{fig:fig3}(d). Grader 2 tended to under-segment the vessel (G1vG2, low DFPD and high DFND scores), while grader 1 tended to over-segment (G2vG1, high DFPD and low DFND scores). As desired, our segmentation tended to be within the region of uncertainty between the two graders (see Figs. \ref{fig:fig3}(c) and \ref{fig:fig3}(d)). Accordingly, the mean Dice score and mean Hausdorff distance of our algorithm against grader 1 (0.917$\pm$0.019, 0.097$\pm$0.019mm) and grader 2 (0.905$\pm$0.018, 0.091$\pm$0.019mm) were better than the inter-grader scores of (0.892$\pm$0.019, 0.105$\pm$0.02mm). \textit{The largest observed Hausdorff distance error of 0.135mm is 6 times smaller than the smallest observed vessel diameter of 0.81mm. Similarly, the mean Hausdorff distance error of 0.094$\pm$0.019mm is $\sim$7 times smaller than smallest observed vessel diameter. This satisfies our goal of sub-mm vessel contour localization.} Tracking was successful as the vessel contours in all sequences were segmented.



\noindent
\textbf{HFUS Results.} To show the generality of our approach to HFUS, we ran our algorithm on 5 HFUS sequences (250 images each), and the corresponding results are shown in Fig. \ref{fig:fig2}(e) and Figs. \ref{fig:fig3}(e)-\ref{fig:fig3}(h). As opposed to UHFUS, lower DFPD and DFND scores were seen with HFUS, meaning a greater consensus in grader annotations (see Figs. \ref{fig:fig3}(g) and \ref{fig:fig3}(h)). Notably, our algorithm still demonstrated the desirable property of final segmentations that lay in the uncertain region of annotation between the two graders. This is supported by comparing the mean Dice score and mean Hausdorff distance of our algorithm against grader 1 (0.915$\pm$0.008, 0.292$\pm$0.023mm) and grader 2 (0.912$\pm$0.021, 0.281$\pm$0.065mm), with the inter-grader scores (0.915$\pm$0.02, 0.273$\pm$0.04mm). To compare against the 0.1mm Mean Absolute Deviation (MAD) error in \cite{Chaniot16}, we also computed the MAD error for HFUS sequences (not shown in Fig. \ref{fig:fig3}). The MAD error of our algorithm against grader 1 was 0.059$\pm$0.021mm, 0.057$\pm$0.024mm against grader 2, and 0.011$\pm$0.003mm between the graders. \textit{Despite the lower pixel resolution ($92.5$\textit{\SI{}{\micro\meter}}) of the HFUS machine used in this work, our MAD errors were $\sim$2$\times$ lower than the state-of-the-art 0.1mm MAD error in \cite{Chaniot16}.} Furthermore, only minor changes in the parameters of the algorithm were required to transfer the methodology to HFUS sequences; namely, the bilateral filter size was 3$\times$3 pixels, ${w}_{o}$=5, and $\Delta{\tau}$=8. No other changes were made to the level set parameters. 

\noindent
\textbf{Performance.} The average run-time on an entry-level NVIDIA GeForce GTX 760 GPU was $19.15$ millisecond per B-scan and $1.915$ seconds per sequence, thus achieving a potential real-time frame rate of $52$ frames per second. The proposed approach is significantly faster than the regular CPU- \cite{Chaniot16}, and real-time CPU- \cite{Abolmaesumi00,Guerrero07,Wang09} and GPU-based approaches in \cite{Smistad16} respectively. Efficient use of CUDA unified memory and CUDA programming contributed to the performance speed-up.

\section{Conclusion and Future Work}
In this paper, a robust system combining the advantages of local phase analysis \cite{Belaid11}, a distance-regularized level set \cite{Li10}, and an Extended Kalman Filter (EKF) \cite{Kalman60} was presented to segment and track vessel contours in UHFUS sequences. The approach, which has also shown applicability to traditional HFUS sequences, was validated by two graders, and it produced similar results as the expert annotations. To the best of our knowledge, this is the first system capable of rapid deformable vessel segmentation and tracking in UHFUS images. Future work is directed towards multi-vessel tracking capabilities. 

\noindent
\textbf{Acknowledgements.} NIH 1R01EY021641, DOD awards W81XWH-14-1-0371 and W81XWH-14-1-0370, NVIDIA Corporation, and Haewon Jeong. 



\begin{thebibliography}{4}
\small
\bibitem{Gorantla11}V. Gorantla et al., ``Acute and Chronic Rejection in Upper Extremity Transplantation: What Have We Learned?", Hand Clinics, 27(4), pp.481-493, (2011).

\bibitem{Abolmaesumi00}P. Abolmaesumi et al., "Real-Time Extraction of Carotid Artery Contours from Ultrasound Images", IEEE Symp. Comp. Med. Sys., pp.181-186, (2000).

\bibitem{Guerrero07}J. Guerrero et al., ``Real-Time
Vessel Segmentation and Tracking for Ultrasound Imaging Applications", IEEE Trans. Med. Imag., 26(8), pp.1079-1090, (2007).

\bibitem{Wang09}D. Wang et al., ``Fully Automated Common Carotid Artery and Internal Jugular Vein Identification and Tracking using B-Mode Ultrasound", IEEE Biomed. Eng., 56(6), pp.1691-1699, (2009).

\bibitem{Smistad16}E. Smistad et al., ``Real-Time Automatic Artery Segmentation, Reconstruction and Registration for Ultrasound-Guided Regional Anaesthesia of the Femoral Nerve", IEEE Trans. Med. Imag., 35(3), pp.752-761, (2016).

\bibitem{Chaniot16}J. Chaniot et al., ``Vessel Segmentation in High-Frequency 2D/3D Ultrasound Images", IEEE Int. Ultrasonics Symp., pp.1-4, (2016).

\bibitem{Tomasi1998}C. Tomasi, ``Bilateral Filtering for Gray and Color Images", ICCV, (1998).

\bibitem{Stetten13}G. Stetten et al., ``Descending Variance Graphs for Segmenting Neurological Structures", Pat. Recog. Neuroimaging (PRNI), pp.174-177, (2013).

\bibitem{Belaid11}D. Boukerroui et al., ``Phase-Based Level Set Segmentation of Ultrasound Images", IEEE Trans. Inf. Technol. Biomed., 15(1), pp.138-147, (2011).


\bibitem{Fitzgibbon95}A. Fitzgibbon, ``A Buyer's Guide to Conic Fitting", BMVC, 2, pp.513-522, (1995).

\bibitem{Li10}C. Li et al., ``Distance Regularized Level Set Evolution and its Application to Image Segmentation", IEEE Trans. Image Process., 19(12), (2010).

\bibitem{Kalman60}R. E. Kalman, ``A New Approach to Linear Filtering and Prediction Problems", J. Fluids Eng., 82(1), pp.35-45, (1960).


\bibitem{Maurer03}C. Maurer, ``A Linear Time Algorithm for Computing Exact Euclidean Distance Transforms of Binary Images in Arbitrary Dimensions," IEEE PAMI, 25(2), (2003).

\end{thebibliography}
\end{document}